\documentclass[11pt,a4paper]{article}
\usepackage{times,latexsym}
\usepackage{url}
\usepackage[T1]{fontenc}

\usepackage[acceptedWithA]{tacl2021v1}

\usepackage{xspace,mfirstuc,tabulary}

\newif\iftaclinstructions
\taclinstructionsfalse 
\iftaclinstructions
\renewcommand{\confidential}{}
\renewcommand{\anonsubtext}{(No author info supplied here, for consistency with
TACL-submission anonymization requirements)}
\newcommand{\instr}
\fi

\iftaclpubformat 

\else

\fi

\usepackage{rotating}
\usepackage{array,booktabs}
\newcolumntype{P}[1]{>{\centering\arraybackslash}p{#1}}
\newcolumntype{M}[1]{>{\arraybackslash}m{#1}}
\usepackage{changepage}
\usepackage{ragged2e} 
\usepackage[most]{tcolorbox}

\usepackage{fancyhdr}
\title{Demographic Biases and Gaps in the Perception of Sexism in Large Language Models}

\author{
 Judith Tavarez-Rodríguez\thanks{~~This work was presented as a poster at the Latin American Meeting in Artificial Intelligence KHIPU 2025, Santiago, Chile, March 10th – 14th 2025, https://khipu.ai/khipu2025/poster-sessions-2025/}~~$^\diamond$ 
  \and
  Fernando Sánchez-Vega$^{\diamond \dagger}$ 
  \and
  A. Pastor López-Monroy$^\diamond$
  \\
  \ \\
  $^\diamond$ Computer Science Department, Mathematics Research Center (CIMAT)\\ Jalisco S/N Valenciana, 36023, Guanajuato, Guanajuato, México
  \\
  $^\dagger$Secretaría de Ciencia, Humanidades, Tecnología e Innovación (SECIHTI),\\ Av. Insurgentes Sur 1582, Col. Crédito Constructor, 03940, CDMX, México\\
  \texttt{\{judith.tavarez, fernando.sanchez, pastor.lopez\}@cimat.mx}
}

\date{}

\begin{document}
\maketitle
\begin{abstract}
The use of Large Language Models (LLMs) has proven to be a tool that could help in the automatic detection of sexism. Previous studies have shown that these models contain biases that do not accurately reflect rea\-lity, especially for minority groups. Despite various efforts to improve the detection of sexist content, this task remains a significant challenge due to its subjective nature and the biases present in automated models.We explore the capabilities of different LLMs to detect sexism in social media text using the EXIST 2024 tweet dataset. It includes anno\-tations from six distinct profiles for each tweet, allowing us to evaluate to what extent LLMs can mimic these groups’ perceptions in sexism detection. Additionally, we analyze the demographic biases present in the models and conduct a statistical analysis to identify which demographic characteristics (age, gender) contribute most effectively to this task.Our results show that, while LLMs can to some extent detect sexism when considering the overall opinion of populations, they do not accurately replicate the diversity of perceptions among different demographic groups. This highlights the need for better-calibrated models that account for the diversity of perspectives across different populations.
\end{abstract}

\section{Introduction}
Sexism has traditionally been defined as a set of attitudes, beliefs, and behaviors that foster gender-based discrimination. Its interpretation, however, differs across theoretical paradigms from disciplines such as philosophy, sociology, and gender studies, some of which emphasize how social class and ethnicity intersect to shape distinct experiences of sexism \cite{hooks2000feminist}.

Within these frameworks, cultural, demographic, and historical factors influence how societies perceive and respond to sexist discourse. In communities with deeply rooted traditions, certain behaviors may not be viewed as offensive, whereas in contexts with stronger debates on gender equality, the same acts may be considered unacceptable. Such diversity highlights the need to account for context-specific features to identify subtle forms of sexism and gauge social tolerance \cite{rosenthal2014polyculturalism, mckerl2007multiculturalism}. We adopt this contextual perspective to enrich our analysis, while acknowledging standpoint epistemology’s view that sexism should be defined through the experiences of marginalized or affected groups \cite{collins2022BlackFeministThought, harding1986cienceQuestioninFeminism, Pohlhaus2002}.

The study of sexism now extends beyond the social sciences into computational approaches. The rise of social media and large-scale data availability has fueled the development of Natural Language Processing (NLP) tools to detect sexist expressions, enabling targeted monitoring and moderation. These methods reveal both global and regional discourse patterns but require cultural adaptation to account for linguistic, demographic, and social diversity.

Recent work has explored whether autonomous systems can detect sexism \cite{rodriguez2021overview}, driven by the urgency to reduce harmful biases and foster inclusivity online \cite{anzovino2018misogyny,lei2024systematic, Gobbo2025misogyny}. Large Language Models (LLMs) have achieved notable success in hate speech and offensive language detection, yet they often inherit and may amplify biases from training data \cite{bolukbasi2016man}, making them both powerful and potentially problematic. Understanding how LLMs address sensitive issues like sexism is critical, as biased classifications can have serious social consequences \cite{elsherief2018hate,vidgen2021directions}.

Despite the growing research on LLM-based sexism detection, few studies examine how perceptions vary across demographic groups \cite{huang-etal-2020-multilingual} and how such variation affects model behavior. The EXIST 2024 dataset \cite{plaza2024overview} offers a benchmark for this purpose, capturing annotations from a demographically diverse pool and thus allowing the study of the phenomenon considering the demographic information of the annotators of the dataset \cite{jimenez2025enhancing}.

This work addresses that remaining gap by exploring whether LLMs can simulate (“profile”) distinct demographic viewpoints in sexism detection and align their responses with the nuanced perspectives reflected in human annotations. We also analyze perspective bias by comparing unconstrained model outputs with those guided by assigned demographic profiles. Our primary goal is to evaluate the extent to which an LLM can reproduce human perceptions of sexism under systematically applied demographic constraints.

Specifically, we pose the following \textbf{research questions}:
\begin{enumerate}
    \item How to induce a profile into an LLM to evaluate its perception of sexism?
    \item In which demographic group is the perception of sexism best replicated by the LLM when assigning a profile?
\end{enumerate}
By focusing on these questions, we address a novel angle that moves beyond mere classification performance and delves into replicating the subjective lens through which different annotators perceive sexism.

The contributions of this work are as follows:
First, we conduct a systematic comparison of multiple prompting templates for assigning demographic profiles to LLMs, identifying the prompt that yields the most statistically reliable profiling.
Second, we quantify the extent to which profiled LLM outputs align with the distribution of annotations from different demographic groups, thereby offering insights into the demographic biases present in these models.

\section{Related Work}\label{sec:related_work}

In recent years, research on sexism detection in online content has advanced notably, with the EXIST (sEXism Identification in Social neTworks) challenge playing a key role in driving the community toward state-of-the-art performance \cite{rodriguez2021overview, rodriguez2022overview, plaza2023overview, plaza2024overview}. Progress has been made both in the technical aspects of sexism detection and in incorporating demographic considerations into annotation \cite{tian2024large, jimenez2024analysis, jimenez2025enhancing}, yet the intersection between these two areas remains underexplored.

A common strategy for probing model bias and introducing controlled demographic context into LLM outputs is to prompt the model to adopt explicit identities such as gender, age, or nationality \cite{gupta2023bias,giorgi-etal-2024-modeling, kim2025exploringpersonadependentllmalignment, liu-etal-2024-evaluating-large, tan-lee-2025-unmasking}. While \citet{zheng-etal-2024-helpful} report that assigning personas does not always improve performance, they also show that factors such as gender, persona type, and domain can influence results. Given that sexism detection is inherently tied to gender, it is crucial to assess whether these findings hold in this specific context.

Combining effective profiling strategies with well-curated datasets offers a promising path for examining the role of cultural factors in shaping LLM behavior. Such an approach not only deepens understanding of how cultural context affects model outcomes \cite{CulturalbiasaculturalalignmentTaoYanandViberg, moralityAksoy2025} but also provides valuable insights into the societal implications of these systems, underscoring the risks of deploying them indiscriminately without considering these influences.

By leveraging the richly annotated and demographically diverse EXIST 2024 dataset, the contributions of our study lie in two principal directions. First, we introduce a statistically supported search to identify prompts that yield a demographic-specific persona. Second, we move beyond uni-variate profiling by analyzing joint demographic configurations (gender, age and nationality), which uncovers compound biases undetectable in isolated factors.

\section{Inducing Profiles to LLMs}\label{sec:induce_profiles}
To address the research question \textit{How to induce a profile into an LLM to evaluate its perception of sexism?} we seek to embed different demographic profiles into LLMs and assess whether a model, when prompted to act as a person with specific demographic traits, generates responses that align with that profile in the context of sexism detection.
\subsection{Experimental Setup}\label{sec:induce_profiles_setup}
For demographic profiling, we evaluate multiple prompt designs, including one of the proposed by \citet{gupta2023bias} (P1 in Table~\ref{tab:profiling_prompts}), as well as hybrid variants that combine manually crafted prompts with Gupta’s template (P2 and P3 in Table~\ref{tab:profiling_prompts}). The profiles are induced using different combinations of gender, age, and country (Table \ref{tab:profile_possible_values}), with the templates adapted to reflect each specific demographic value. The \textit{\{profile\}} part of every template is replaced by the following according to the characteristics listed:
\begin{center}
    \begin{itemize}
        \item \textbf{Gender}: \{gender\}
        \item \textbf{Age}: \{age\}-year-old person
        \item \textbf{Gender and Age}: \{age\}-year-old \{gender\}
        \item \textbf{Gender, Age and Country}: \{age\}-year-old \{gender\} from \{country\}
    \end{itemize}
\end{center}

\begin{table}[ht]
\centering
  \begin{tabular}{lM{6cm}}
    ID & Profiling Prompt \\
    \hline
    P1 & \textit{You are a \{profile\}. Your responses should closely mirror the knowledge and abilities of this persona.} \\
    \hline
    P2 & \textit{Act as if you are a \{profile\}. Your responses should closely mirror the knowledge and abilities of this persona.} \\ \hline
    P3 & \textit{You are a \{profile\}. Interpret the tweet from this perspective and apply the classification.}\\ \hline
  \end{tabular}
   \caption{Profiling prompts to evaluate the best form to induce a demographic profile into the LLMs. }
  \label{tab:profiling_prompts}
\end{table}

\begin{table}[ht]
\begin{tabular}{M{2.5cm}M{4.2cm}}
Profiling \newline Demographics  & Possible Values  \\ \hline
Gender                  & Female, Male    \\ \hline
Age                     & 18-22, 23-45, 46+  \\ \hline
Gender and Age          & Female \& 18-22,\newline Female \& 23-45,\newline Female \& 46+,\newline Male \& 18-22,\newline Male \& 23-45, \newline Male \& 46+  \\ \hline
Gender, Age and Country & Female \& 18-22 \& Italy, \newline Female \& 23-45 \& Mexico, \newline Female \& 46+ \& Canada,\newline Male \& 18-22 \& Spain,\newline Male \& 23-45 \& Spain,\newline Male \& 46+ \& Germany \\ \hline                         
\end{tabular}
\caption{Demographic characteristics used for profiling the LLMs. Note: possible values of countries in this table are just illustrative.}
\label{tab:profile_possible_values}
\end{table}

For each profiling prompt, we carried out the sexism detection task by instructing the LLM to classify the tweets in the user’s request as either sexist or non-sexist. For this purpose, we employed the following \textbf{user-instruction-prompt}:

\begin{quote}
\ttfamily
"**Instructions for Classification:**\\
    - **YES**: Classify the
tweet as YES if it exhibits sexism directly, describes a sexist scenario, or criticizes sexist behavior.\\
-**NO**: Classify the tweet as NO if it does not show prejudice against, undermine, or discriminate against women.

**Tweet**: "
\end{quote}

This prompt, originally proposed in \citet{tavarez2024better}, was developed through a prompt engineering process based on the EXIST dataset annotation guidelines. We replicated the experiments reported in that work using a different LLM, and our findings were consistent with the authors’ results.

\paragraph{Data and Models} We employed the TRAIN and DEV partitions of the EXIST dataset to induce demographic profiles in LLMs and examine whether their outputs aligned with the assigned profiles. The dataset, annotated by six human profiles, offers a solid reference for validating whether induced profiles reproduce consistent patterns.\\
The model used in our experiments was Meta/Llama-3.1-8B-Instruct \cite{grattafiori2024llama}, with all prompts issued in English. We tested two temperature settings: 0, to produce highly deterministic responses and assess the model’s ability to consistently follow a profile, and 1, the default in many APIs, to reflect how profile induction might perform under typical commercial configurations.

\paragraph{Procedure} Each complete run involved classifying all 7,958 tweets six times, corresponding to the six demographic profiles that replicate the EXIST dataset structure. For example, in gender profiling, three classifications were performed assuming a female role and three assuming a male role. For every demographic variant (Table \ref{tab:profile_possible_values}), we carried out two full runs—one for each temperature setting—resulting in a total of eight full runs per profiling prompt. This setup provides more reliable estimates for identifying the profiling prompt that most effectively induces demographic roles.\\
Model outputs (YES or NO) were extracted using regular expressions, yielding six binary predictions per tweet. These were aggregated into a soft prediction for the positive class, computed as the average number of YES responses:
$$soft\_prediction = \frac{\#YES}{6}$$
The consistency of the induced profiles was assessed by comparing these soft predictions with the dataset’s soft golden labels using Pearson correlation \cite{pearson1896vii}. Higher correlation values indicate that the induced profiles more closely replicate the annotators’ demographic perspectives.

\subsection{Results and Discussion}
\begin{table*}[ht]
\centering
  \begin{tabular}{M{6cm}P{1.2cm}P{1.2cm}P{2cm}P{2cm}P{1cm}}
    \hline
    Profiling Prompt & Gender & Age & Gender-Age & Gender-Age-Country & Avg \\
    \hline
    \textit{You are a \{profile\}. Your responses should closely mirror the knowledge and abilities of this persona.} &   0.4134&	\textbf{0.5401}&	\textbf{0.5215}&	\textbf{0.5552}&	\textbf{0.5076} \\
    \hline
    \textit{Act as if you are a \{profile\}. Your responses should closely mirror the knowledge and abilities of this persona.} &0.4379&	0.5269	&0.5194	&0.5286&	0.5032\\ \hline
    \textit{You are a \{profile\}. Interpret the tweet from this perspective and apply the classification.}& \textbf{0.4590}	&0.4556	&0.4864&	0.3572&	0.4396
\\ \hline
  \end{tabular}
   \caption{Correlation by Profiling Demographics: Average of correlations between soft predictions and soft golden labels from 8 complete runs (Avg column) performed at temperatures 0 and 1 for each profiling prompt. Model: Llama-3.1-8B-Instruct.}
  \label{tab:results_profiling_prompts}
\end{table*}

Table \ref{tab:results_profiling_prompts} summarizes the results of the experiments on inducing demographic profiles in an LLM. Of all the prompts evaluated, the approach proposed by \citet{gupta2023bias} delivered the highest overall performance, leading in three out of the four demographic profiling combinations. The highest correlations with human annotations were observed for the Gender–Age–Country combination, while profiling solely by gender produced the lowest performance across all demographics.

Interestingly, the prompt that achieved the best results for gender profiling proved significantly less effective for other demographics, showing a drop of nearly seven points—an appreciable difference. Another noteworthy observation is that the first two prompts outperform all others by more than six points. The second prompt is a slight variation of the first, differing only in its initial profiling sentence. This pattern suggests that the effectiveness of Gupta’s prompt may stem largely from its second sentence: "Your responses should closely mirror the knowledge and abilities of this persona."

\section{Comparisons of Profiled LLMs and Human Annotators}\label{sec:comparisons_llm_vs_human}

We compared profiled LLMs with the human annotations in the EXIST dataset to determine which demographic group’s perception of sexism is most accurately reproduced by an LLM when given the corresponding profile.

\subsection{Experimental Setup}
The sexism detection task followed the procedure outlined in Section~\ref{sec:induce_profiles}, using the best-performing demographic profile from Table~\ref{tab:results_profiling_prompts} alongside the user-instruction prompt described in Section~\ref{sec:induce_profiles_setup}.\\
For the comparison between LLMs with demographic profiling and human annotations, we generated “synthetic human” annotations aligned with the demographic characteristics of the actual annotators. As in Section~\ref{sec:induce_profiles}, the model was queried six times per tweet (one full run), varying the demographic attributes. Two full runs were conducted for each profiling variant in Table~\ref{tab:profile_possible_values}—gender, age, gender+age, and gender+age+country—once at temperature 0 and once at temperature 1, yielding a total of eight full runs.

\paragraph{Models and Procedure}
All prompt requests were issued in English using Llama-3.1-8B-Instruct and Gemini-1.5-Flash-8B, and the model outputs for the sexism detection task were extracted via regular expressions.\\
For demographic profiling, we computed the proportion of “YES” votes as soft predictions, both for the consensus of all six synthetic annotations and separately by demographic group (female, male, 18–22, 23–45, 46+). These soft predictions were then compared to the corresponding soft labels from the human annotations using Pearson correlation.\\
The neutral profile experiments followed the same procedure, except that only a single soft prediction per tweet—representing the consensus of six queries—was generated and compared against both the overall and demographic soft labels.

\subsection{Results and Discussion}
Table \ref{tab:comparisons_llms_human} reports the correlation values between profiled models and different human demographic groups.

For Llama-3.1-8B, the highest correlation with the consensus of the six annotators occurs when profiling with the Gender–Age–Country combination, except for the 46+ group, where age-only profiling yields better results. In general, age-based profiling produces higher correlations than gender-based profiling, while combining age and gender lowers the correlation compared to age alone. When using gender-only profiling, the model aligns more closely with male annotators; however, adding age reverses this trend, increasing alignment with female annotators. Overall, Llama-3.1-8B tends to reflect women’s perception of sexism more closely, although gender-only profiling shifts it toward men. By age, the model correlates highest with the 23–45 group and lowest with the 18–22 group.

For Gemini, gender-only profiling produces the strongest correlation with the consensus. Within age groups, combining gender and age results in higher correlations than age alone. In most profiling setups (except age-only), the model’s perception of sexism aligns more strongly with female demographics, whereas in the age-only configuration, the highest correlation is again with the 23–45 group.

\begin{table*}[ht]
\centering
  \begin{tabular}{llllllll}
    Model & Profile & All & F & M & 18-22 & 23-45 & 46+ \\
    \hline
    Llama-3.1-8B & Gender &0.4134 &0.3449 &\underline{0.3811} & & & \\
    Llama-3.1-8B & Age &0.5401 & & &0.415 &\underline{0.4572} &\textbf{0.4272} \\
    Llama-3.1-8B & Gender-Age &0.5215 &\underline{0.4808}&0.4596 &0.3814 &0.4472 &0.4084 \\
    Llama-3.1-8B & Gender-Age-Country &\textbf{0.5552} &\underline{\textbf{0.5061}} &\textbf{0.4802} &\textbf{0.431} &\textbf{0.4748} &0.4125\\
    Llama-3.1-8B & Neutral &0.5874&	\underline{0.5463}&	0.5197&	0.4773&	0.4991&	0.4951 \\ \hline
    Gemini 1.5 flash 8B & Gender &\textbf{0.6227} &\underline{\textbf{0.5617}} &\textbf{0.5423} & & & \\
    Gemini 1.5 flash 8B & Age & 0.5884& & &0.472 &\underline{0.4859} &0.45 \\
    Gemini 1.5 flash 8B & Gender-Age &0.5884 &\underline{0.5376} &0.5199 &\textbf{0.4729} &\textbf{0.4959} &\textbf{0.4639}\\
    Gemini 1.5 flash 8B & Gender-Age-Country & 0.5522&\underline{0.5021} &0.4772 &0.4362 &0.4666 &0.4226 \\
    Gemini 1.5 flash 8B & Neutral &0.6096& \underline{0.563}& 0.5433 &0.4967& 0.5171& 0.5132 \\ \hline
  \end{tabular}
   \caption{Pearson correlation. \textbf{Bold}: Best result per profiling (vertical), except neutral profile. \underline{Underlined}: Best result per demographic (horizontal) except overall consensus.}
  \label{tab:comparisons_llms_human}
\end{table*}

\section{Conclusions and Future Work}
This study enhances the process of selecting an effective prompt for inducing demographic profiles in LLMs by broadening the scope of experiments, thereby providing a stronger basis for choosing a profiling strategy. Among the evaluated prompts, the one proposed by \citet{gupta2023bias} achieved the highest average performance in sexism detection when demographics were assigned, with its second sentence appearing particularly influential for profiling success.\\
Comparisons between model outputs and human annotations revealed that different profiling configurations produce varying perceptions of sexism, though these do not fully align with the views of their corresponding human demographic groups. Overall, the models tended to align more closely with female demographics, suggesting that their perception of sexism is more strongly influenced by women’s viewpoints. Future work could investigate the linguistic distinctions between the best overall profiling prompt and the one that performed best for gender-specific profiling, to better understand the notable divergence in gender-based results.\\
Additional research could also examine the relationship between the 23–45 age group and each model’s perception of sexism, as well as explore how modifying system roles or user instructions based on the language of the tweet might alter the observed patterns.\\
Under the configurations tested, our findings indicate that LLMs still fall short in replicating the full diversity of perceptions across demographic groups, highlighting the need for more precisely calibrated models that better reflect varied cultural and demographic perspectives.

\section*{Limitations}

This study has several limitations that affect its generalizability to the wider population.
First, in the EXIST dataset, certain regions or countries are represented by only a small number of annotators, meaning that the data for some demographics—particularly those involving geographic location—relies on very limited perspectives. Furthermore, as noted in the dataset documentation, there is no representation of indigenous communities, tribal groups, or diverse racial profiles.
Second, all analyses were performed using smaller model versions (8B parameters), leaving the behavior and implications for much larger models unexplored. In addition, every prompt was issued in English, even when evaluating Spanish tweets, which could have influenced model outputs.
Finally, model refusals were classified as sexist cases based on the nature of the refusals; however, this assumption warrants further investigation.

\section*{Ethics Statement}

For clarity, we use the term “perception” to describe emergent properties of model training. This does not imply any form of intentionality or cognition on the part of the model. The analyses and conclusions presented in this work are strictly confined to the text domain and the metrics applied.
Furthermore, we acknowledge that sexism encompasses multiple theoretical currents, many of which are beyond the scope of this study. Here, the term “sexist” is used specifically in reference to a sociological perspective in which sexism is understood as a phenomenon shaped by individuals’ perceptions. We recognize that framing sexism solely in terms of individual perception risks obscuring the lived realities of vulnerable groups. Therefore, future work should prioritize incorporating the perspectives and experiences of those most affected by sexism, ensuring that these voices are not rendered invisible within the analysis.

\bibliographystyle{acl_natbib}

\onecolumn

\end{document}